\def\year{2021}\relax
\documentclass[letterpaper]{article} 
\usepackage{aaai21}  
\usepackage{times}  
\usepackage{helvet} 
\usepackage{courier}  
\usepackage[hyphens]{url}  
\usepackage{graphicx} 
\usepackage{natbib}  
\usepackage{caption} 
\usepackage{graphicx}
\usepackage{amsmath}
\usepackage{amsfonts}
\usepackage{xspace}
\usepackage{multirow}
\usepackage{longtable}
\usepackage{supertabular}
\usepackage{paralist}
\usepackage{cite}
\usepackage{booktabs}
\usepackage{pifont}
\usepackage{amssymb}
\newcommand{\cmark}{\ding{51}}
\newcommand{\xmark}{\ding{55}}
\usepackage{array}
\usepackage{color}
\definecolor{darkblue}{rgb}{0,0,0.5}
\definecolor{lightpink}{rgb}{1,0.71,0.75}
\usepackage{algorithm,algpseudocode}

\newcommand{\loss}{\ensuremath{\mathcal{L}}}
\newcommand{\xvec}{\ensuremath{\mathbf{x}}}
\newcommand{\yvec}{\ensuremath{\mathbf{y}}}
\newcommand{\zvec}{\ensuremath{\mathbf{z}}}
\newcommand{\hvec}{\ensuremath{\mathbf{h}}}

\newcommand{\vvec}{\ensuremath{\mathbf{v}}}

\newcommand{\method}{LUT\xspace}

\title{Listen, Understand and Translate: \\ Triple Supervision Decouples End-to-end Speech-to-text Translation}

\author {
    Qianqian Dong, \textsuperscript{\rm 1,2}\thanks{The work was done while QD was interning at ByteDance AI Lab.}
    Rong Ye, \textsuperscript{\rm 3}
    Mingxuan Wang, \textsuperscript{\rm 3}
    Hao Zhou, \textsuperscript{\rm 3}
    Shuang Xu, \textsuperscript{\rm 1}
    Bo Xu, \textsuperscript{\rm 1,2}
    Lei Li \textsuperscript{\rm 3}\\
}
\affiliations {
    \textsuperscript{\rm 1} Institute of Automation, Chinese Academy of Sciences (CASIA), Beijing, China \\
    \textsuperscript{\rm 2} School of Artificial Intelligence, University of Chinese Academy of Sciences, Beijing, China \\
    \textsuperscript{\rm 3} ByteDance AI Lab, China \\
    \{dongqianqian2016, shuang.xu, xubo\}@ia.ac.cn, 
    \{yerong, wangmingxuan.89, zhouhao.nlp, lileilab\}@bytedance.com
}

\begin{document}

\maketitle

\begin{abstract}
An end-to-end speech-to-text translation (ST) takes audio in a source language and outputs the text in a target language.
Existing methods are limited by the amount of parallel corpus. 
Can we build a system to fully utilize signals in a parallel ST corpus? 
We are inspired by human understanding system which is composed of auditory perception and cognitive processing. 
In this paper, we propose \textbf{L}isten-\textbf{U}nderstand-\textbf{T}ranslate, (\method), a unified framework with triple supervision signals to decouple the end-to-end speech-to-text translation task.
\method is able to guide the acoustic encoder to extract as much information from the auditory input. 
In addition, \method  utilizes a pre-trained BERT model to enforce the upper encoder to produce as much semantic information as possible, without extra data. 
We perform experiments on a diverse set of speech translation benchmarks, including Librispeech English-French, IWSLT English-German and TED English-Chinese. 
Our results demonstrate \method achieves the state-of-the-art performance, outperforming previous methods. 
The code is available at \url{https://github.com/dqqcasia/st}.
\end{abstract}

\section{Introduction}
\label{sec:intro}
\begin{figure*}[ht]
\centering
\includegraphics[width=0.95\textwidth]{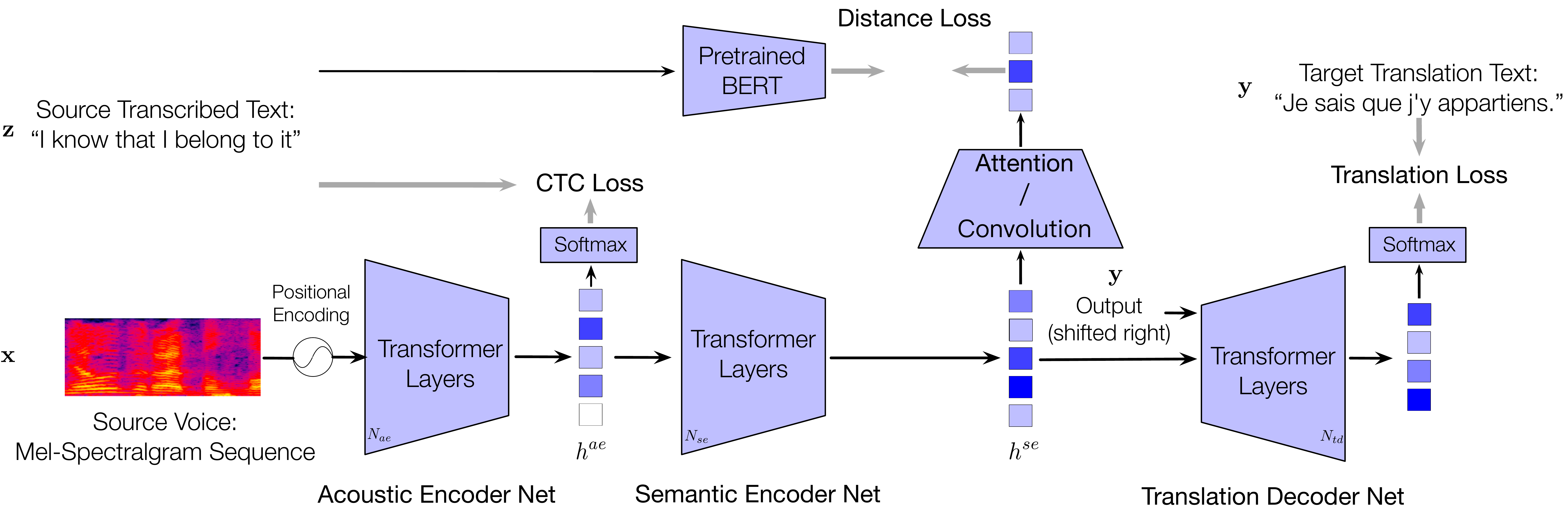}
\caption{The architecture of \method. It contains three modules, an acoustic encoder, a semantic encoder, and a translation decoder.}
\label{fig:model}
\end{figure*}

Processing audio in one language and translating it into another language has been requested in many applications. 
Traditional speech translation (ST) systems are cascaded by connecting separately trained automatic speech recognition (ASR), spoken language normalization processing and machine translation (MT) subsystems~\citep{sperber2017neural,wang2018semi,dong2019adapting,sperber2019self,zhang2019lattice,beck2019neural,cheng2019breaking}. 
However, such cascaded ST systems have drawbacks including higher latency, larger memory footprint, and potential error propagation in the subsystems. 
In contrast, an end-to-end ST system has a single unified model, which is beneficial in deployment. 
While very promising, existing end-to-end ST models still cannot outperform cascaded systems in terms of translation accuracy. 

Cascaded ST systems usually have intermediate stages which extract acoustic features and source-text semantic features, before translating to the target text, like humans with perception systems and cognitive systems to process different information. 
Ideally, a neural encoder-decoder network should also benefit from imitating these intermediate steps. 
The challenges are: 
\begin{inparaenum}[a)]
    \item there is no sufficient supervision to guide the internals of an encoder-decoder to process the audio input and obtain acoustic and semantic information properly.
    \item the training corpus for ST with pairs of source audio and target text is much smaller than those typically used for ASR and MT.
\end{inparaenum}
Previous works attempt to relieve these challenges using pre-training and fine-tuning approaches.
They usually initialize the ST model with the encoder trained on ASR data to mimic the speech recognizing process and then fine-tune on a speech translation dataset to make the cross-lingual translation.
However, pre-training and fine-tuning are still not sufficient enough to train an effective ST system, for the following reasons: \begin{inparaenum}[\it a)]
\item the encoder for speech recognition is mainly used to extract acoustic information, while the ST model requires to encode both acoustic and semantic information. 
\item previous studies \citep{battenberg2017exploring} have proved that the learned alignments between input and output units in the ASR models are local and monotonic, which is not conducive to modeling long-distance dependencies for translation models.
\item the gap of length between the input audio signals (typically $\sim 1000$ frames) and target sentences (typically $\sim 20$ tokens) renders the association from the encoder to decoder difficult to learn. 
\end{inparaenum}

Based on the above analysis, we explore decoupled model structure, \method, with an acoustic encoder (to \textbf{L}isten), a semantic encoder (to \textbf{U}nderstand), and a translation decoder (to \textbf{T}ranslate) to imitate the intermediate steps for effective end-to-end speech translation.
In addition to the widely-used translation loss with cross-entropy, we exploit two additional auxiliary supervising signals. 
We utilize the monotonic alignment for transcriptions to ensure the acoustic encoder capture necessary acoustic information from the input audio spectrum sequence. 
In this way, the local relations among the nearby audio frames are preserved. 
We utilize the pre-trained embedding, to guide the semantic encoder to capture a proper semantic representation. 
Notice that neither of the two auxiliary supervision is required during the inference, and therefore our method is efficient. 

The contributions of the paper are as follows:
\begin{inparaenum}[\it 1)]
\item We design \method, a unified framework augmented with additional components and supervisions to decouple and guide the speech translation.
\item Our proposed method can extract semantic knowledge from the pre-trained language model and utilize external ASR corpus to enhance acoustic modeling more effectively benefiting from the flexibly designed structure.
\item
We conduct experiments and do analysis on three mainstream speech translation datasets, LibriSpeech (English-French), IWSLT2018 (English-German) and TED (English-Chinese), to verify the effectiveness of our model.
\end{inparaenum}

\section{Methodology}
\label{sec:approach}
In this section, we illustrate how we design the speech-to-text translation model, \method, whose architecture allows a flexible configuration of the backbone network structure in each module. One can freely choose convolutional layers, recurrent neural networks, or \emph{Transformer} network as the main building structure.  Figure~\ref{fig:model} illustrates the overall architecture of the \method, using \emph{Transformer}~\cite{vaswani2017attention} as the backbone network. Our proposed \method consists of three modules: 
\begin{inparaenum}[\it a)]
    \item an \emph{acoustic encoder} network that encodes the audio input sequence into hidden features corresponding to the source text;
    \item a \emph{semantic encoder} network that extracts hidden semantic representation for translation, which behaves like a normal machine translation encoder;
    \item a \emph{translation decoder} network that outputs sentence tokens in the target language. 
\end{inparaenum}
Notice an input sequence typically has a length of more than $1000$, while a target sentence has tens of tokens. 
\paragraph{Problem Formulation}
The training corpus for speech translation contains speech-transcription-translation triples, denoted as $\mathcal{S} = \{(\xvec, \zvec,\yvec)\}$. 
Specially, $\xvec=(x_1,...,x_{T_x})$ is a sequence of acoustic features. $\zvec=(z_1,...,z_{T_{z}})$ and $\yvec=(y_1,...,y_{T_{y}})$ represents the corresponding text sequence in source language and target language, respectively. Meanwhile, $\mathcal{A}=\{(\xvec',\zvec')\}$ represents the external ASR corpus. Usually, the amount of ST corpus is much smaller than that of ASR, i.e. $|\mathcal{S}| \ll |\mathcal{A}|$. 

\subsection{Acoustic Encoder}
\label{sec:at}
The acoustic encoder of \method takes the input of low-level audio features and outputs a series of vectors corresponding to the transcribed text in the source language. 
The original audio signal is transformed into mel-frequency cepstrum~\citep{mermelstein1976distance}, which is the standard preprocessing in speech recognition.
The sequence of frames ($\xvec$) are processed by a feed-forward linear layer, and $N_{ae}$ layers of \emph{Transformer} sub-network, which includes a multi-head attention layer, a feed-forward layer, normalization layers, and residual connections. 
The output of acoustic encoder is denoted as $\hvec^{ae}$.
They are further projected linearly with a softmax layer to obtain auxiliary output probability $p$ for each token in the vocabulary. 
Note here the vocabulary is augmented with one extra blank symbol ``\textvisiblespace'' (``$<$blk$>$" in the vocabulary). 
We employ the alignment-free Connectionist Temporal Classification (CTC)~\citep{graves2006connectionist} algorithm to align the acoustic encoder output and the expected supervision sequence $\zvec$.

\paragraph{CTC Loss} Given the ground truth transcribed token sequence $\zvec$, there can be multiple raw predicted label sequences from the acoustic encoder. 
Let $g$ denote the mapping from the raw label sequence to the ground truth, which is based on a deterministic rule by removing the blank symbols and consecutive duplicate tokens. 
For example, $g(aa\textvisiblespace a b\textvisiblespace) = g(a \textvisiblespace a bb\textvisiblespace) = aab$.
$\mathcal{B}^{-1}(\zvec)$ is the set of all raw label sequences corresponding to the ground truth transcription.
Then the conditional probability of a ground truth token sequence  $\zvec$ can be modeled by marginalizing over all raw label sequences:
\begin{equation}
P(\zvec|\xvec)=\sum_{\pi \in\mathcal{B}^{-1}(\zvec)}P(\pi|\xvec)
\end{equation}
Where each raw label probability $p(\pi_t|\xvec)$ for a sequence $\pi$ is calculated from the acoustic encoder using the following equation:
\begin{equation}
P(\pi|\xvec)=\prod_{t=1}^{T_x} p(\pi_t|\xvec)
\end{equation}
Finally, the acoustic encoder loss is defined as 
\begin{equation}
\loss_{ae}(\theta;\xvec,\zvec)=-\log P(\zvec|\xvec)
\label{eq:acoustic_loss}
\end{equation}

\subsection{Semantic Encoder}
\label{sec:se}
The second module of \method is motivated by the commonly used encoder for a neural machine translation (NMT) model. 
\method's semantic encoder aims to extract semantic and contextual information for translation. 
However, unlike the encoder in the NMT model taking the input of source sentence tokens, \method's semantic encoder takes the hidden representation $\hvec_{ae}$ computed from the acoustic encoder as the input.
Since we do not have explicit supervision of the semantic representation, we utilize a pre-trained BERT~\cite{devlin2019bert} model to calculate sentence embeddings for the source sentence $\zvec$ and then further employ these embeddings to supervise the training of this encoder module. 
This approach of self-supervision is advantageous because it enables training using a very large independent monolingual corpus in the source language. 
The semantic encoder contains $N_{se}$ \emph{Transformer} layers at the core and then connects to two branches. 
The output of this module is denoted as $\hvec^{se}$. 
One branch is to compute an overall semantic vector of the input, marked as ``Seq-level Distance". It is realized using a 2D convolutional layer to reduce dimension, a normalization layer, and an average pooling layer to shrink the vectors into one. 
The output of this branch is denoted as $v_{0}^{se}$, which is a single vector. 
This is to be compared with the class-label representation $h_c^{\text{BERT}}$ calculated by a BERT model. 
Another branch is aimed to match the semantic representation of the transcribed source sentence, marked as ``Word-level Distance". 
This branch is connected to an auxiliary layer to calculate the length-synchronized semantic representation, note as $v_{1}^{se}$, a sequence of vectors of the size $T_z$, equivalent to that of the transcribed source sentence. 
To this end, we first use a separately pre-trained BERT model to calculate the sentence embedding vectors, excluding the class-label vector $h_c^\text{BERT}$. 
These vectors are organized into $T_z$ time steps, denoted as $\hvec^\text{BERT}$.
Suppose the $N_{se}$-layer transformer outputs a sequence of vectors at length $T_x$, denoted as $\vvec=\hvec^{se}$. 
Note each of these vectors are split into $J$ heads, i.e. $\hvec^\text{BERT}=(\hvec_1^\text{BERT}, \dots, \hvec_J^\text{BERT})$ and $\vvec=(\vvec_1, \dots, \vvec_J)$. 
These BERT vectors are used as queries to compute the attention weights for the branch input hidden vectors. 
\begin{align}
    \text{head}_i = \text{Attn}(\hvec_i^\text{BERT} W_i^Q, \vvec_i W_i^K, \vvec_i W_i^V)
\end{align}
Where the $W_i^Q$,$W_i^K$,$W_i^V$ are parameters for the attention of $i$-th head. The attention is calculated by scaled dot-product layer, as follows:
\begin{equation}
\text{Attn}(Q,K,V)=\text{Softmax}(\frac{Q\cdot K^{T}}{\sqrt{d_k}})V
\end{equation}
where $d_k$ is the dimension of the key $K$.
With this layer, the output can be reduced to the same length as source text by concatenating the heads.  
\begin{equation}
\vvec_1^{se} = \text{Concat}(\text{head}_1,\dots, \text{head}_J)
\end{equation}
Finally, the distance loss is defined as the mean-squared error loss (MSE) between the calculated hidden representations and the BERT embeddings.
\begin{equation}
\loss_{se}(\theta;\xvec,\zvec) =
\begin{cases}
     \text{MSE}(v_0^{se}-h_c^{\text{BERT}}), & \text{Seq-level} \\
     \text{MSE}(\vvec_1^{se} - \hvec^{\text{BERT}}), & \text{Word-level}
\end{cases}
\label{eq:semantic_encoder_loss}
\end{equation}

The key insight of our formulation is that the semantic encoder needs to behave like a text encoder of the NMT model, with only source language text data in the training. 
The distance loss (Equation~\ref{eq:semantic_encoder_loss}) ensures that the semantic encoder could produce similar semantic embeddings close to the BERT representation trained on a separate large text corpus.
During the inference, the output of this module is $\hvec^{se}$, therefore no additional source transcription text is needed and the BERT calculation is saved.

\subsection{Translation Decoder}
As with the normal machine translation model, our proposed \method uses $N_{td}$ layers of \emph{Transformer} network as the decoder. 
Additional attention from the decoder to the semantic encoder output $\hvec^{se}$ is added.
The cross entropy loss is defined in Equation~\ref{eq:decoder_loss}.
\begin{equation}
\loss_{td}(\theta;\xvec,\yvec)=-\sum_{i=1}^{T_y}\log p_{\theta}(y_i|y_{\text{\textless} i}, \hvec^{se})
\label{eq:decoder_loss}
\end{equation}
Where the decoder probability $p_{\theta}$ is calculated from the final softmax layer based on the output of the decoder. 

\paragraph{Training}
The overall objective function for end-to-end training is the weighted sum for three supervision modules:
\begin{equation}
\begin{aligned}
&\loss(\theta;\xvec,\zvec,\yvec) \\
&=\alpha\loss_{ae}(\theta;\xvec,\zvec)+\beta\loss_{se}(\theta;\xvec,\zvec)+\gamma\loss_{td}(\theta;\xvec,\yvec)
\label{eq:overall_loss}
\end{aligned}
\end{equation}
Where $\theta$ is the model parameter, and $\alpha$, $\beta$ and $\gamma$ are hyper-parameters to balance among the acoustic encoder loss $\loss_{ae}$, the semantic encoder loss $\loss_{se}$, and the translation decoder loss $\loss_{td}$.

\begin{algorithm}[t]
\caption{Semi-supervised Training Strategy}
\label{algorithm_expanded}
\begin{algorithmic}[1] 
    \While{not converged}
        \State \textbf{Step 1}: sample a training batch from $\{(\xvec,\zvec)\}$ (or $\mathcal{A}$ in expanded setting) to optimize $\loss_{ae}$ and $\loss_{se}$.
        \State \textbf{Step 2}: sample a training batch from $\mathcal{S}$ to optimize the total loss in Equation~\ref{eq:overall_loss}.
    \EndWhile
\end{algorithmic}
\end{algorithm}

We have also used the semi-supervised training strategy (Algorithm \ref{algorithm_expanded}) to take advantages of the parallel \verb|<speech, transcription>| supervisions and external ASR resources. 
In this way, the end-to-end ST model can avoid the complicated process of pre-training and then fine-tuning, while being able to better converge and maintain acoustic modeling.

\section{Experiments}
\label{sec:exps}
\subsection{Data}
\label{sec:data}

We conduct experiments on three popular publicly available datasets, including Augmented LibriSpeech English-French dataset~\citep{kocabiyikoglu2018augmenting}, IWSLT201 English-German dataset~\citep{jan2018iwslt} and TED English-Chinese dataset~\citep{liu2019end}. 

\paragraph{Augmented LibriSpeech Dataset}
Augmented LibriSpeech is built by automatically aligning e-books in French with English utterances of LibriSpeech. The dataset includes four types of information: English speech signal, English transcription, French text translations from the alignment of e-books with augmented references via Google Translate. Following the previous work~\citep{liu2019end}, we also conduct experiments on the 100 hours clean train set for training, with 2 hours development set and 4 hours test set, corresponding to 47271, 1071, and 2048 utterances respectively. 

\paragraph{IWSLT2018 English-German Dataset}
IWSLT2018 English-German is the KIT end-to-end speech translation corpus, which is built automatically by aligning English audios with SRT transcripts for English and German from lectures online. 
The raw data, including long wave files, English transcriptions, and the corresponding German translations, is segmented into chunks with the attached timestamps. 
It should be noted that some transcriptions are not aligned with their corresponding audio well. Noisy data is harmful to models' performance, which can be avoided by data filtering, re-alignment, and re-segmentation~\citep{liu2018ustc}. In this paper, the original data is used directly as training data to verify our method, with a size of 272 hours and 171121 segmentations. 
We use \textit{dev2010} as validation set and \textit{tst2013} as test set, corresponding to 653 and 793 utterances respectively~\footnote{We use the data segmentation in the $iwslt-corpus/parallel$ folder.}.

\paragraph{TED English-Chinese Dataset}
English-Chinese TED is crawled from TED website\footnote{\url{https://www.ted.com}} and released by~\citep{liu2019end} as a benchmark for speech translation from English audio to Chinese text. Following the previous work~\citep{liu2019end}, we use dev2010 as development set and tst2015 as test set. The raw long audio is segmented based on timestamps for complete semantic information. Finally, we get 524 hours train set, 1.5 hours validation set and 2.5 hours test set, corresponding to 308,660, 835, 1223 utterances respectively.

\paragraph{LIUM2 Dataset}
We use LIUM2 as the external ASR parallel corpus ($\in \mathcal{A}$) used in the expanded experimental setting for broad reproducibility. LIUM2~\citep{rousseau2014enhancing} is composed of segments of public talks extracted from the TED lecture website with 207 hours of speech data. Speed perturbation is performed on the raw signals with speed factors 0.9 and 1.1.

\paragraph{Data Preprocessing}
Following the previous work~\citep{liu2019end}, we use acoustic features that are 80 dimensional log-Mel filterbanks. The features are extracted with a step size of 10ms and a window size of 25ms and extended with mean subtraction and variance normalization. The features are stacked with 5 frames to the right and downsampled to a 30ms frame rate. 
For target language text data, 
we lowercase all the texts, tokenize and apply normalize punctuations with the Moses scripts\footnote{\url{https://github.com/moses-smt/mosesdecoder}}. 
For source language text data, we lowercase all the texts, tokenize and remove the punctuation to make the data more consistent with the output of ASR.
We apply BPE\footnote{\url{https://github.com/rsennrich/subword-nmt}} on the combination of source and target text to obtain shared subword units. The number of merge operations in BPE for ASR and MT systems is set to 8k and 30k, respectively. For strategies using BERT features, we apply the same pre-processing tool as BERT does to text data for ST models and regenerate the vocabulary. For English-French and English-German corpora, we report case-insensitive BLEU scores by \texttt{multi-bleu.pl}\footnote{\url{https://github.com/moses-smt/mosesdecoder/scripts/generic/multi-bleu.perl}} script for the evaluation of ST and MT tasks. And for English-Chinese corpus, we report character-level BLEU scores. We use word error rates (WER) to evaluate ASR tasks. 

\subsection{Baselines}
We conduct our experiments in different settings.

\paragraph{Base Setting with only Speech-translation Data} 
Our main purpose is to compare our method with conventional end-to-end speech translation models. In the setting, the training data is restricted to only the triple data. 

\paragraph{Expanded Setting with External Data}
In the context of expanded setting, ~\citet{bahar2019using} apply the SpecAugment~\citep{park2019specaugment} on Librispeech English-French ST task, which uses a total of 236 hours of speech for ASR pre-training. ~\citet{inaguma2019multilingual} combine three ST datasets of 472 hours training data to train a multilingual ST model for both Librispeech English-French ST task and IWSLT2013 English-German ST task. And \citet{wang2019bridging} introduce an additional 207 hours ASR corpus and 40M parallel data from WMT18 to enhance the ST performance. 
We mainly explored additional ASR data in this work.

\begin{table*}[t]
\centering
\small
\begin{tabular}{lcccc} 
\toprule
 Method & \shortstack{Enc Pre-train\\(speech data)} & \shortstack{Dec Pre-train\\(text data)}  & greedy & beam \\ 
\midrule
\textbf{MT system} & & &    \\
Transformer MT~\citep{liu2019end} & - & - &21.35 & 22.91 \\
\midrule
\textbf{Base ST setting} & & &    \\
LSTM ST~\citep{berard2018end} & \xmark &\xmark & 12.30 & 12.90 \\
~~+pre-train+multitask~\citep{berard2018end} & \cmark & \cmark    &12.60 & 13.40 \\
LSTM ST+pre-train~\citep{inaguma2020espnet} & \cmark & \cmark    &-& 16.68\\
Transformer+pre-train~\citep{liu2019end} & \cmark & \cmark  &13.89& 14.30\\
~~+knowledge distillation~\citep{liu2019end} &\cmark & \cmark &14.96& 17.02\\
TCEN-LSTM~\citep{wang2019bridging}& \cmark & \cmark  &-& 17.05\\
Transformer+ASR pre-train~\citep{wang2020curriculum} &\cmark &\xmark & -  &15.97\\
Transformer+curriculum pre-train~\citep{wang2020curriculum} &\cmark &\xmark & - &17.66\\
\method &\xmark & \xmark & \textbf{16.70} &\textbf{17.75}\\
\midrule 
\textbf{Expanded ST setting}& &  &  \\
LSTM+pre-train+SpecAugment~\citep{bahar2019using} & \cmark (236h)&\cmark &-& 17.00\\
Multilingual ST+PT~\citep{inaguma2019multilingual}  & \cmark(472h)&\xmark &-& 17.60\\
Transformer+ASR pre-train~\citep{wang2020curriculum} &\cmark(960h) & \xmark&  - &16.90\\
Transformer+curriculum pre-train~\citep{wang2020curriculum} &\cmark(960h) &\xmark & - &18.01\\
\method &\cmark(207h) & \xmark&\textbf{17.55} &\textbf{18.34}\\
\bottomrule
\end{tabular}
\caption{Performance on Augmented Librispeech English-French test set. \method achieves the best performance on both the base ST and the expanded settings. }
\label{enfr}
\end{table*}

\paragraph{MT System}
The input of the MT system is the manual transcribed text. The result of MT system can be regarded as the upper bound of ST models.

\subsection{Details of the Model and Experiments}
\label{detail}
For ST tasks, we use a similar hyper-parameter setting with the \emph{Transformer} base model \citep{vaswani2017attention} for the stack of \emph{Transformer} layers, in which we set the hidden size $d_{model}=768$ ($512$ for English-Chinese TED set) to match the output of BERT. We experiment with the officially released BERT ($H=768$) on English-French and English-German and BERT ($H=512$) on English-Chinese. Learning from speech-transformer~\citep{dong2018speech}, one third of the layer is used for the decoder ($N_{td}=4$).
 $N_{ae}$ and $N_{se}$ are both set to 4 for our best performance. For ASR and MT tasks, the standard \emph{Transformer} base model is adopted.
 All samples are batched together with 20000-frame features by approximate feature sequence length during training. 
 We train our models on 1 NVIDIA V100 GPU with a maximum number of training steps 400k, using Adam optimizer~\cite{kingma2014adam} with $\beta_1=0.9$, $\beta_2=0.999$, $\epsilon=1e^{-8}$, and warmup-decay learning rate schedule with peak lr = $4e^{-4}$, 25k warmup steps, 0.5 decay rate and 50k decay steps. 
 We use a greedy search and beam search (as default) with a max beam size of 8 for our experimental settings. The maximum decoding length for ASR and ST (MT) is set to 200 and 250, respectively. The hyper-parameters in Equation \ref{eq:overall_loss}, $\alpha$, $\beta$ and $\gamma$ are set to 0.5, 0.05, 0.45 (details in Table \ref{tab:parameters}). For experiments in the base setting, the ST model is trained from scratch. For experiments in the expanded setting, the ST model is trained as the following two steps: 
 \begin{inparaenum}[\it a)]
 \item pre-training the acoustic encoder with CTC loss with $(\xvec',\zvec') \in \mathcal{A}$, as Equation \ref{eq:acoustic_loss}.
 \item fine-tuning the overall ST model with $(\xvec, \zvec, \yvec) \in \mathcal{S}$ and $(\xvec',\zvec') \in \mathcal{A}$, as Algorithm \ref{algorithm_expanded}.
 \end{inparaenum} 
SpecAugment strategy~\citep{park2019specaugment} is adopted to avoid overfitting with frequency masking (F = 30, mF = 2) and time masking (T = 40, mT = 2).
The final model is averaged on the last 10 checkpoints.

\section{Results}
\label{sec:results}

\begin{table*}[htb]
\centering
\small
\begin{tabular}{p{6.8cm}p{1.5cm}<{\centering}p{1.5cm}<{\centering}p{0.8cm}<{\centering}} 
\toprule
 Method & \shortstack{Enc Pre-train\\(speech data)} & \shortstack{Dec Pre-train\\(text data)} &  tst2013\\
\midrule
\textbf{MT system} & & &    \\
RNN MT~\citep{inaguma2020espnet} & - & - & 24.90 \\
\midrule
\textbf{Base ST setting} & & & \\
ESPnet~\citep{inaguma2020espnet} &\xmark &\xmark & 12.50\\
~~+enc pre-train&\cmark &\xmark &  13.12\\
~~+enc dec pre-train&\cmark &\cmark &  13.54\\
Transformer+ASR pre-train~\citep{wang2020curriculum}&\cmark &\xmark & 15.35\\
~~+curriculum pre-train~\citep{wang2020curriculum}&\cmark & \xmark& 16.27\\
\method  & \xmark & \xmark&    \textbf{16.35}\\
\midrule 
\textbf{Expanded ST setting}& & &  \\
Multilingual ST~\citep{inaguma2019multilingual}&\cmark(472h) &\xmark &14.60\\
CL-fast*~\citep{kano2018structured}&\cmark(479h) &\xmark & 14.33 \\
TCEN-LSTM~\citep{wang2019bridging} &\cmark(479h) &\cmark(40M) &  17.67\\
Transformer+curriculum pre-train~\citep{wang2020curriculum} &\cmark(479h) &\cmark(4M) &  18.15 \\
\method &\cmark(207h) & \xmark&   \textbf{18.59}\\
\bottomrule
\end{tabular}
\caption{Performance on IWSLT2018 English-German test set. *: re-implemented by~\citet{wang2020curriculum}. \method achieves the best performance. }
\label{ende}
\end{table*}

\begin{table*}[ht]
\centering
\small
\begin{tabular}{lccc} 
\toprule
 Method & \shortstack{Enc Pre-train\\(speech data)} & \shortstack{Dec Pre-train\\(text data)}  & BLEU \\  
\midrule
\textbf{MT system} & & &    \\
Transformer MT~\citep{liu2019end} & - & - & 27.08\\ 
\midrule
\textbf{Base setting} & & &    \\
Transformer+pre-train~\citep{liu2019end} & \cmark & \cmark & 16.80 \\
~~+knowledge distillation~\citep{liu2019end}& \cmark &\cmark & 19.55 \\
Multi-task+pre-train*~\citep{inaguma2019multilingual}(re-implemented) & \cmark & \xmark& 20.45 \\
\method & \xmark& \xmark  &\textbf{20.84}\\
\bottomrule
\end{tabular}
\caption{Performance on TED English-Chinese test set. \method achieves the best performance.}
\label{enzh}
\end{table*}

\subsection{Main Results}
\label{sec:res}
\paragraph{Librispeech English-French}
For En-Fr experiments, we compare the performance with existing end-to-end methods in Table \ref{enfr}. Clearly, \method outperforms the previous best results in base setting and expanded setting respectively. 
Specifically, in the base setting, the model we propose outperforms ESPnet, which is equipped with both a well-trained encoder and decoder. 
We also achieve better results than a knowledge distillation augmented baseline in which an MT model is introduced to teach the ST model~\citep{liu2019end}. Different from previous work, our work focuses on reducing the modeling burdens of the encoder by suggesting that auxilliary supervision signals make it easier to learn both the acoustic and semantic information. This proposal promises great potential for the application of the double supervised encoder. Compared to~\citet{wang2019bridging, wang2020curriculum} which also includes tandem encoders, \method is simple and flexible, without the need to introduce additional computational cost for data preprocessing (such as training the noisy MT model or the alignment model). Simple yet effective, \method achieves the best performance in this benchmark dataset in terms of BLEU.

\paragraph{IWSLT2018 English-German}
For En-De experiments, we compare the performance with existing end-to-end methods in Table \ref{ende}.
Unlike that of Librispeech English-French, this dataset is noisy, and the transcriptions do not align well with the corresponding audios. As a result, there is a wide gap between the performance of the end to end ST and the upper bound of the ST. Overall, our method outperforms ESPnet on tst2013 by 2.81 BLEU in the base setting and has an advantage of 0.44 BLEU compared with~\citet{wang2020curriculum} in the expanded setting. 
To be notice, \method does not need any pretraining tricks, and achieves the state-of-the-art performance in the base setting.
This trend is consistent with that on the Librispeech dataset.

\paragraph{TED English-Chinese}
For En-Zh experiments, we compared the performance with existing end-to-end methods in Table \ref{enzh}. Under the base setting, \method exceeded the Transformer-based ST model augmented by knowledge distillation with 1.29 BLEU, proving the validity of our method.

\paragraph{Comparison with Cascaded Baselines}
Table~\ref{cascaded} shows the comparison with cascaded ST systems on Augmented Librispeech En-Fr test set, IWSLT2018 En-De tst2013 set and TED En-Zh test set. 
For a fair comparison, we do the experiments on the base settings of English-French/German/Chinese translation. Results show that \method either receives the equivalent performance or outperforms with cascaded methods on three datasets, thus displaying great potential for the end-to-end approach. This indicates our flexible structure can make good use of additional ASR corpus and learn valuable linguistic knowledge. 

\begin{table}[ht]
\centering
\small
\begin{tabular}{cll} 
\toprule
 &Method & BLEU \\  
\midrule
\multirow{2}*{\shortstack{$En$$\rightarrow$$Fr$}} & cascaded & 17.58 \\
&\method & 17.75\\
\midrule
\multirow{2}*{\shortstack{$En$$\rightarrow$$De$}} & cascaded & 15.38 \\
&\method & 16.35\\
\midrule
\multirow{2}*{\shortstack{$En$$\rightarrow$$Zh$}} & cascaded & 21.36 \\
&\method & 20.84\\
\bottomrule
\end{tabular}
\caption{\method versus cascaded systems on Augmented Librispeech En-Fr test set, IWSLT2018 En-De tst2013 set and TED En-Zh test set. ``cascaded" systems consist of separate ASR and MT models trained independently.}
\label{cascaded}
\end{table}

\subsection{Analysis}
\label{sec:ablation}
\paragraph{Effects of Auxiliary Supervision}
We first study the effects of two auxiliary supervision for \method. The results in Table \ref{ablation} show that all the auxiliary supervision indicate positive results that can be superimposed. Models that use supervision only from the acoustic encoder can be regarded as a method of multi-task learning, which has a significant performance improvement compared to the model of direct pre-training and fine-tuning (seen in Table \ref{enfr}). This reflects the catastrophic forgetting problem that occurs in the sequential transfer learning based on the pre-training method. 

\begin{table}[ht]
\centering
\small
\begin{tabular}{lcc}   
\toprule
 & Dev Bleu & Test Bleu \\  
\midrule        
  \method  & \textbf{18.51} & \textbf{17.75} \\
  ~~~ w/o Semantic Encoder Loss  & 17.72  & 16.81 \\
  ~~~ w/o Acoustic Encoder Loss *  & 16.91 & 15.48 \\
  ~~~ w/o Acoustic Encoder Loss & 12.05 & 11.24 \\
\bottomrule
\end{tabular}
\caption{Ablation study on En-Fr validation and test set. ``*" means using ASR pre-training as initialization.}
\label{ablation}
\end{table} 

\paragraph{Balance of Acoustic and Semantic Modeling}
Experimental results, shown in Table \ref{layers}, prove that the performance is better when the two modules are balanced.
In order to determine which module has a more significant impact on performance, we conducted experiments on the layer number allocation of the two modules, in which the total number of layers of the acoustic encoder and semantic encoder is fixed, and the number of layers for one module is adjusted from 2 to 6. As the number of layers decreases, the two modules will both result in worse performance degradation, thus explaining that using enough layers to extract acoustic features and encode semantic representation is 
equally essential to the speech translation model.

\begin{table}[ht]
\centering
\small
\begin{tabular}{cccc}   
\toprule
$N_{ae}$&$N_{se}$&  Dev BLEU & Test BLEU \\  
\midrule        
  2 & 6  & 14.81  & 13.09 \\
  3 & 5  & 17.01 & 15.50 \\
  4 & 4  & \textbf{17.93} & \textbf{16.70}\\
  5 & 3  & 17.07 & 16.21 \\
  6 & 2  & 16.47 & 15.49\\
\bottomrule
\end{tabular}
\caption{
Performance on En-Fr corpus: \method with varying layers in its acoustic encoder ($N_{ae}$) and semantic encoder ($N_{se}$). Greedy decoding is employed.
}
\label{layers}
\end{table}

\paragraph{Sequence-level Distance Versus Word-level Distance}

For this part, we conduct experiments with different branches described in previous sections. We conduct an experimental comparison of the performance differences caused by the pre-training features extracted by different layers of BERT for semantic encoder's supervision. 
We finally adopt the pre-trained features from the last layer of BERT as our default setting. The results, as shown in Table \ref{distilling}, prove that the word-level distance benefit more from the BERT pre-trained features because of its finer and grainer regulation. 

\begin{table}[ht]
\centering
\small
\begin{tabular}{ccc}   
\toprule
 & Dev BLEU & Test BLEU \\  
\midrule        
  Seq-level Distance& 17.64  & 16.61 \\
  Word-level Distance& \textbf{17.93} & \textbf{16.70} \\
\bottomrule
\end{tabular}
\caption{
Performance on En-Fr corpus: \method with different losses for semantic encoder. ``Seq-level'' and ``Word-level'' losses are described in Eq.~\eqref{eq:semantic_encoder_loss}.
Greedy decoding is employed.
}
\label{distilling}
\end{table}

\paragraph{Effect of Hyper-parameters in the Loss}

We vary the loss weights of the three terms in loss function, i.e. $\alpha$, $\beta$ and $\gamma$ in Equation~\ref{eq:overall_loss}. The results are listed in Table~\ref{tab:parameters}. 
For different settings, we want to emphasize the influence of :
\begin{itemize}
    \item Translation loss $\loss_{td}$ (No. IV);
    \item CTC loss $\loss_{ae}$ (No. VI);
    \item Distance loss $\loss_{se}$ (No. V);
    \item Both CTC loss $\loss_{ae}$ and translation loss $\loss_{td}$ (No. I \& II), 
    \item All of them (No. III).
\end{itemize}
The performances do not vary much ($\pm 0.5$ BLEU) after convergence. 
We also find that distance loss $\loss_{se}$ is the easiest term to converge while CTC loss $\loss_{ae}$ is the hardest during the training. 

\begin{table}[]
    \centering
    \begin{tabular}{c|ccc|cc}
    \toprule
    No. & $\alpha$ & $\beta$ & $\gamma$ & dev-BLEU & tst-BLEU \\
    \hline
    I & 0.50 & 0.05 & 0.45 & 18.61 & 17.55 \\
    II & 0.40 & 0.20 & 0.40 & 18.56 & 17.00 \\
    III & 0.30 & 0.40 & 0.30 & 18.70 & 17.34 \\
    IV & 0.20 & 0.05 & 0.75 & 18.75 & 17.41 \\
    V & 0.20 & 0.60 & 0.20 & 18.94 &  17.17\\
    VI & 0.80 & 0.05 & 0.15 & 18.38 & 17.31 \\
    \bottomrule
    \end{tabular}
    \caption{Ablations on the loss weight on Augmented Librispeech En-Fr dataset, with greedy decoding under the expanded setting.}
    \label{tab:parameters}
\end{table}

\paragraph{Shallower or Deeper Encoders}
In Table~\ref{layers}, we have concluded that both acoustic and semantic encoders are equally important to the entire system, while in this section, we explore the performance of the deeper or shallower balanced encoders (Table~\ref{tab:layers}). The model with shallower encoders performs worse due to the limited expressive capacity. 
While the encoders go deeper, the 5-layer-encoder model translates slightly better. However the 6-layer-encoder model performs worse than the 4-layer one. This is probably due to the overfitting problem, also mentioned in \citep{wang2020curriculum}. In addition, we guess that a deeper model with a smaller hidden size can achieve better performance, which is reserved for future exploration.

\begin{table}[h]
    \centering
    \begin{tabular}{cccc}
    \toprule
    $N_{ae}$ & $N_{se}$ & $N_{td}$ & test BLEU \\
    \hline
    6 & 6 & 6 & $<5$ \\
    6 & 6 & 4 & 17.15 \\
    5 & 5 & 4 & \textbf{17.58} \\
    4 & 4 & 4 & 17.55 \\
    3 & 3 & 4 & 16.39 \\
    2 & 2 & 4 & 14.13 \\
    \bottomrule
    \end{tabular}
    \caption{Ablations on the number of encoder and decoder layers on Augmented Librispeech dataset, with greedy decoding under the expanded setting.}
    \label{tab:layers}
\end{table}

\paragraph{Acoustic or Semantic}

In Table \ref{probing_task}, we design auxiliary probing tasks to further analyze the learned representation ~\citep{lugosch2019speech}.   
\textbf{SpeakerVer} is designed to identify the speaker, therefore it benefits more from acoustic information. \textbf{IntentIde} is focused on intention recognition, so it needs more linguistic knowledge.
We use the Fluent Speech Commands dataset~\citep{lugosch2019speech} for experiments which contains 30,043 utterances, 97 speakers, and 31 intents. 
The utterances are randomly divided into train, valid, and test splits in such a way that each split contains all possible wordings for each intent and all speakers. 
For the train split, we extract the hidden output of each layer of our well-trained \method encoder, average and freeze it, followed by a fully connected layer. We then fine-tune for 20,000 steps on the two probing tasks respectively. We report the accuracy of the test split.
It can be seen that during the modeling process of ST, acoustic information is modeled at low-level layers and semantic information is captured at high-level layers.

\begin{table}[h]
\centering
\small
\begin{tabular}{ccc} 
\toprule
 &SpeakerVer & IntentIde\\
\midrule
AE Output $\hvec^{ae}$ & \textbf{97.6} &91.0\\
SE output  $\hvec^{se}$ &46.3 &\textbf{93.1}\\
\bottomrule
\end{tabular}
\caption{Classification accuracy on speaker verification and intent identification, using \method's acoustic encoder output ($\hvec^{ae}$) and semantic encoder output ($\hvec^{se}$).}
\label{probing_task}
\end{table}

\paragraph{Model Sizes for ST Systems}
We make a detailed comparison between the performance of end-to-end systems and cascaded systems with different model parameter sizes. Compared with the original pipeline system, the pipeline (small) system is a cascade of speech recognition models and machine translation models with halved layers. Results in Table \ref{model_size} prove that end-to-end models have advantages in balancing performance and parameter size of the system.

\begin{table}[ht]
\centering
\small
\begin{tabular}{lccccccc} 
\toprule
& \multicolumn{3}{c}{En-Fr} & \multirow{2}{*}{Parameter}\\
\cline{2-4}
 & ASR$\downarrow$ & MT$\uparrow$ & ST$\uparrow$ \\  
\hline
 Pipeline (small) & 21.3 & 19.5 &15.9 & $\approx 129M$\\
 Pipeline & 16.6 & 20.98 & 16.38 &$\approx 209M$\\
\method & - & - & $\textbf{16.70}$ &  $\approx 144M$ \\
\bottomrule
\end{tabular}
\caption{Performance with greedy search for ASR, MT and ST tasks on En-Fr test set with different model sizes. Pipeline: consists of independently trained ASR and MT systems (see details in the experimental settings).}
\label{model_size}
\end{table}

\begin{figure}[h]
\centering
\includegraphics[scale=0.42]{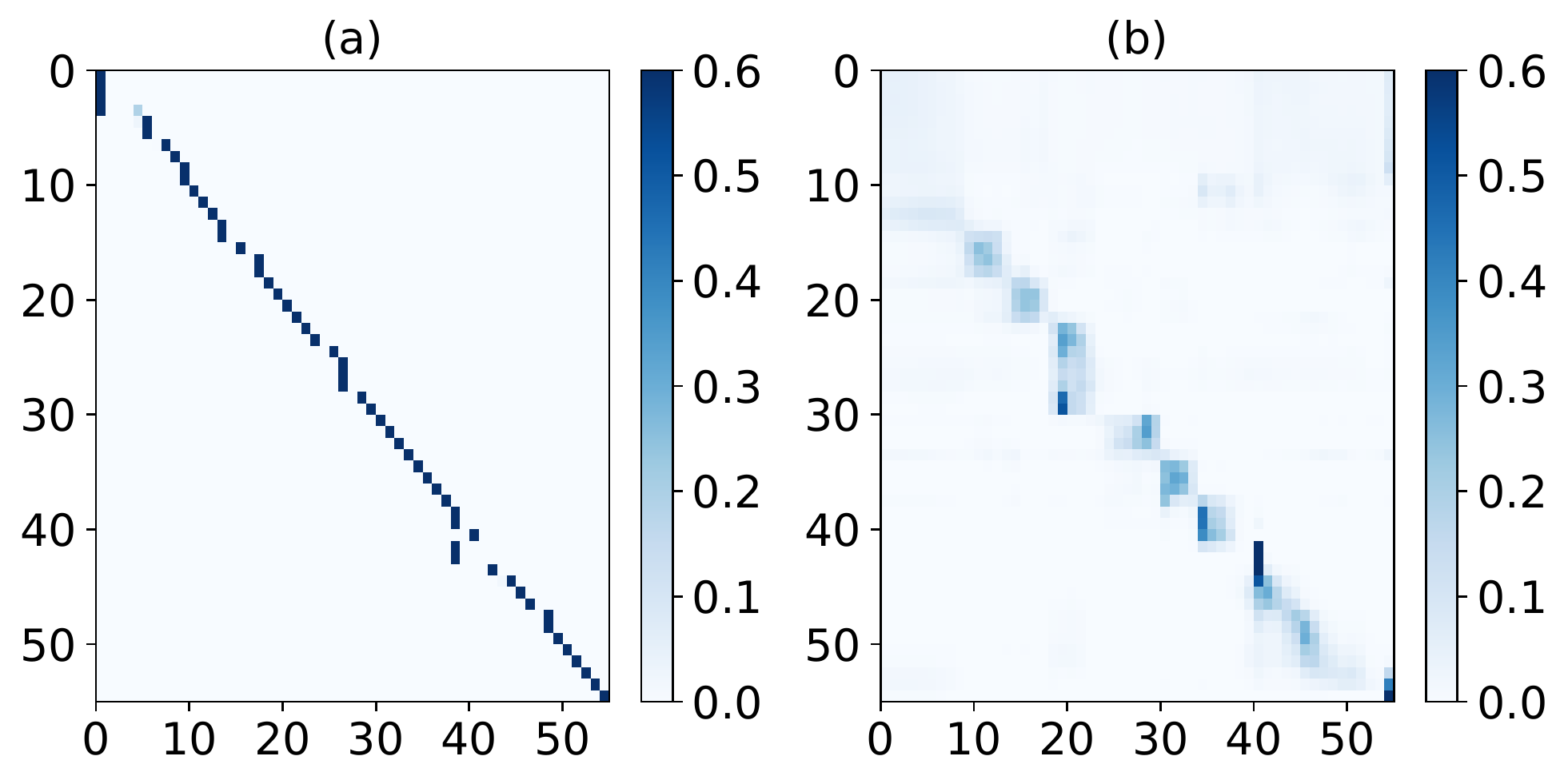}
\caption{The visualization of attention for different module layers. (a), (b) visualize the attention of the last layer of acoustic encoder and the first layer of semantic encoder respectively. Both the horizontal and vertical coordinates represent the same sequence of speech frames.}
\label{attention}
\end{figure}

\noindent\textbf{Attention Visualization}~
We analyze the learned representation through visualizations of the acoustic and semantic modeling's attention between layers. Figure \ref{attention} shows an example of the distribution of attention weights. The attention of acoustic modeling is local and monotonous from the first layer to the fourth layer, matching the behavior of ASR. The attention of semantic encoder gradually tends to be smoothed out across the global context, which is beneficial to modeling semantic information. The observation is in line with our hypothesis. 

\paragraph{Correlation Analysis}
\begin{figure}[h]
\centering
\includegraphics[width=0.48\textwidth]{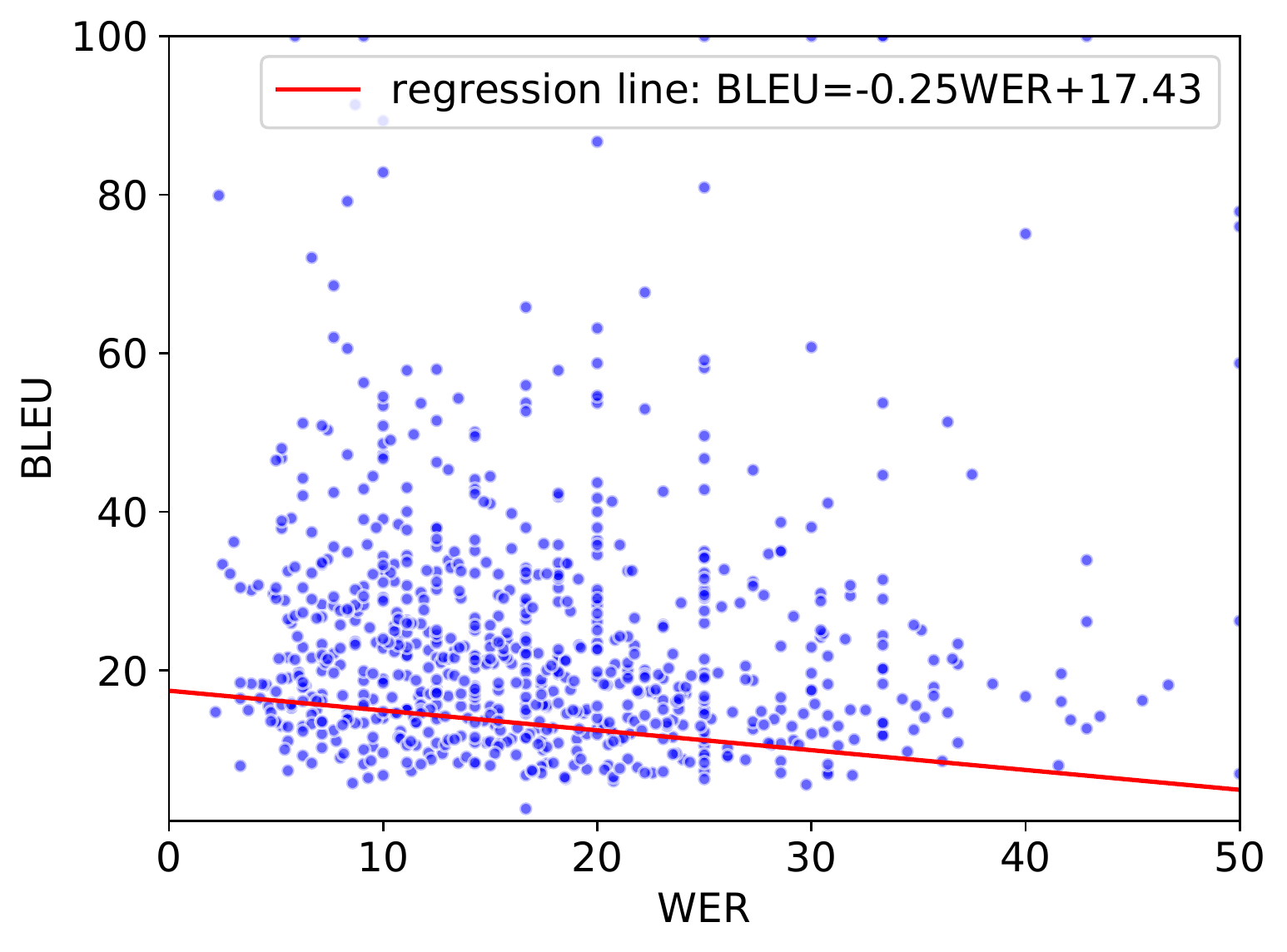}
\caption{Relationship between WER and BLEU on En-Fr test set.}
\label{scatter}
\end{figure}
The quality of the hidden state obtained in our encoder depends largely on the accuracy of the acoustic modeling. Using the CTC loss function introduced in the acoustic encoder, we can also predict recognition results while predicting translation results. We can diagnose whether the wrong prediction for translation is caused by the wrong acoustic modeling in this way. We use samples from the test set to analyze the relationship between translation quality and acoustic modeling, which are evaluated by BLEU and WER respectively. We draw scatter plots of WER and BLEU on the test set, as can be seen in Figure \ref{scatter}. It can be seen that samples with a higher WER can usually obtain a translation result with a lower BLEU. 
Statistically, the Pearson correlation coefficient between BLEU and WER is $-0.205 < 0$ (with p-value = $2e^{-16} <<0.05$), which indicates the significant negative relation between them. 
At the same time, a minority of samples with a higher WER can obtain translation results with a higher BLEU, 
which indicates that our ST model has a certain degree of robustness to recognition errors.

\paragraph{Case Study}
Table~\ref{Case} shows our case study analysis, proving that the end-to-end speech translation system can alleviate the problem of error propagation caused by upstream speech recognition errors. \method can obtain the intermediate results of speech recognition by the way of CTC decoding, so it can perform a certain degree of interpretable diagnosis on the translation results. Benefiting from the ability of the end-to-end system to directly obtain the original audio information, our method is fault-tolerant in the case of incorrect recognition, missing recognition, repeated recognition, and so on during the acoustic modeling. 
\begin{table}[!ht]
    \centering
    \small
    \begin{tabular}{lp{2.25in}}
    \toprule
      \textbf{Speech} \#1 & 766-144485-0090.wav\\
      \textbf{Transcription} & \\
      \textit{reference} & it was mister jack \underline{\textit{maldon}}  \\
     \textit{hypothesis} &it was mister jack \textbf{mal}\\
      \textbf{Translation} & \\
      \textit{reference} &  c'était m. jack \underline{\textit{maldon}} \\
      \textit{hypothesis} & c'était m. jack \underline{\textit{maldon}} \\
      \textbf{Phenomenon} & Incorrect Recognition \\
     \midrule
     \textbf{Speech} \#2 & 1257-122442-0101.wav\\
      \textbf{Transcription} & \\
      \textit{reference} &cried the \underline{\textit{old}} soldier \\
     \textit{hypothesis} &cried the soldier\\
      \textbf{Translation} & \\
      \textit{reference} &s'écria le \underline{\textit{vieux}} soldat, \\
      \textit{hypothesis}&s'écria le \underline{\textit{vieux}} soldat, \\
      \textbf{Phenomenon} & Missing Recognition\\
      \midrule
     \textbf{Speech} \#3 & 1184-121026-0000.wav\\
      \textbf{Transcription} & \\
      \textit{reference} &chapter \underline{\textit{seventeen}} the abbes chamber\\
     \textit{hypothesis} &chapter \underline{\textit{seventeen}} \textbf{teen} the abbey chamber\\
      \textbf{Translation} & \\
      \textit{reference} & chapitre \underline{\textit{xvii}} la chambre de l'abbé. \\
      \textit{hypothesis}  & chapitre \underline{\textit{xvii}} la chambre de l'abbé.\\
      \textbf{Phenomenon} & Repeated Recognition\\
     \bottomrule
    \end{tabular}
    \caption{Examples of transcription and translation on En-Fr test set generated by \method. The underlined text means the desired output, and the bold text represents the incorrect predictions.}
    \label{Case}
\end{table}

\section{Related Work}
\label{sec:related}
\noindent\textbf{End-to-end ST}~
Previous works \citep{berard2016listen,duong2016attentional} have proved the potential for end-to-end ST, which has attracted intensive attentions~\citep{salesky2018towards,di2019adapting,bahar2019comparative,di2019enhancing,inaguma2020espnet}. It's proved that pre-training~\citep{weiss2017sequence,berard2018end,bansal2018pre,stoian2020analyzing} and multi-task learning~\citep{vydana2020jointly} can significantly improve the performance. Two-pass decoding~\citep{sung2019towards} and attention-passing~\citep{anastasopoulos2018tied,sperber2019attention} techniques are proposed to handle deeper relationships and alleviate error propagation in end-to-end models. Data augmentation techniques~\citep{jia2019leveraging,pino2019harnessing,bahar2019using} are proposed to utilize ASR and MT corpora to generate fake data. Pre-training has brought great gains to end-to-end models, such as knowledge distillation~\citep{liu2019end}, modality agnostic meta-learning~\citep{indurthi2019data}, curriculum learning~\citep{kano2018structured,wang2020curriculum} and so on~\citep{wang2019bridging}. \citet{liu2019synchronous,liu2020low} optimize the decoding strategy to achieve low-latency end-to-end ST.  
\citet{salesky2019exploring} explored additional features to enhance end-to-end models. 

\noindent \textbf{Knowledge Distillation for MT}~Representations learned by pre-training have been widely applied in the field of machine translation. There're different modeling granularities for representation learning, which can be utilized by fine-tuning, freezing, or distillation. Many knowledge distillation methods have been extended to transfer the rich knowledge using teacher-student architecture~\citep{gou2020knowledge}. ~\citet{kim2016sequence} extended word-level knowledge distillation into sequence-level knowledge distillation for directing the sequence distribution of the student model.
Sequence-level knowledge distillation was further explained from the perspective of data augmentation and regularization in~\citet{gordon2019explaining}. 
~\citet{zhou2019understanding} studied how knowledge distillation affects the non-autoregressive MT models by empirical analysis.
~\citet{lample2019cross,edunov2019pre} feeded the last layer of ELMo or BERT to the encoder of MT model for learning better representation.
~\citet{yang2020towards} firstly leveraged asymptotic distillation to transfer the pre-training information to MT model.

\section{Conclusion}
\label{sec:conclusion}
In this paper, we propose \method, a novel and unified training framework to decouple the end-to-end speech translation task into three phases: ``Listen", ``Understand" and ``Translate", supervised by the acoustic, semantic, linguistic supervision, respectively. We empirically demonstrate the effectiveness of our approach as compared to previous methods on three benchmark datasets, and empirical analysis suggests that \method is capable of capturing all acoustic, semantic and cross-lingual information properly.  

\section{Acknowledgements}
\label{sec:acknow}
We would like to thank the anonymous reviewers for their valuable comments. We would also like to
thank Chengqi Zhao, Cheng Yi and Chi Han for their useful suggestions and helps. This work was supported by the Key Programs of the Chinese Academy of Sciences under Grant No.ZDBS-SSW-JSC006-2. 

\bibliography{aaai21}
\end{document}